# Event and Activity Recognition in Video Surveillance for Cyber-Physical Systems

Swarnabja Bhaumik, Prithwish Jana and Partha Pratim Mohanta

**Abstract**

In this chapter, we aim to aid the development of Cyber-Physical Systems (CPS) in automated understanding of events and activities in various applications of video-surveillance. These events are mostly captured by drones, CCTVs or novice and unskilled individuals on low-end devices. Being unconstrained in nature, these videos are immensely challenging due to a number of quality factors. We present an extensive account of the various approaches taken to solve the problem over the years. This ranges from methods as early as Structure from Motion (SFM) based approaches to recent solution frameworks involving deep neural networks. We show that the long-term motion patterns alone play a pivotal role in the task of recognizing an event. Consequently each video is significantly represented by a fixed number of key-frames using a graph-based approach. Only the temporal features are exploited using a hybrid Convolutional Neural Network (CNN) + Recurrent Neural Network (RNN) architecture. The results we obtain are encouraging as they outperform standard temporal CNNs and are at par with those using spatial information along with motion cues. Further exploring multistream models, we conceive a multi-tier fusion strategy for the spatial and temporal wings of a network. A consolidated representation of the respective individual prediction vectors on video and frame levels is obtained using a biased conflation technique. The fusion strategy endows us with greater rise in precision on each stage as compared to the state-of-the-art methods, and thus a powerful consensus is achieved in classification. Results are recorded on four benchmark datasets widely used in the domain of action recognition, namely Columbia Consumer Videos (CCV), Human Motion Database (HMDB), UCF-101 and Kodak's Consumer Video (KCV). It is inferable that focusing on better classification of the video sequences certainly leads to robust actuation of a system designed for event surveillance and object cum activity tracking.

## 1 Introduction

Images and motion pictures were until a point of time, a manifestation of lucid art and complex ideas that only humans could appreciate and analyze, owing to their priceless gifts of vision and cognition. The fact that even computers could comprehend and obtain a semantic and high-level understanding of what digital images and videos had to offer was a dream unrealized before the advent of the computer vision. At present, computer vision is an extensive field of study, and an active domain of research. In this chapter, we deal with the problem of event recognition. The goal of event recognition is to understand the various diverse *events* such as Birthday, Graduation, Wedding-Reception, and *activities* like Dancing, Horse-Riding, Walking and so on. These events and activities are mostly captured by drones, CCTVs or novice and unskilled individuals on low-end capturing devices. As a result, these videos are often found to have inadequate brightness, jitter and, are temporally redundant on a number of occa-

S. Bhaumik
Department of Computer Science and Engineering, Meghnad Saha Institute of Technology, Kolkata, India
e-mail: swarnabjazq22@gmail.com

P. Jana
Department of Computer Science and Engineering, Indian Institute of Technology, Kharagpur, India
e-mail: pjana@ieee.org

P. P. Mohanta (✉)
Electronics and Communication Sciences Unit, Indian Statistical Institute, Kolkata, India
e-mail: partha.p.mohanta@gmail.com

51

sions. Camera movement and cluttered background are also prevalent. These videos are termed as 'unconstrained' video. The present work focuses on the classification of unconstrained videos abundantly available either in the prerecorded form in online video-sharing platforms such as YouTube and Facebook, or in real-time. Being unconstrained in nature, these videos have often posed a challenge to researchers. On the other hand, the rising demand for retrieval of high-level information based on the content of the video, the problem of event recognition has garnered much interest. This brings us at a confluence of the broad streams of machine vision and artificial intelligence (with sub-fields being machine learning and deep learning) to deal the problem in real-life scenarios. With a better and enforced automated understanding of surrounding events, a number of significant problems such as unmanned video surveillance and tracking in various domains including sports, defense, and other related areas, can also be addressed. The challenge is to represent each of these complex events or activities with their underlying unique structure/pattern. With the prevalence and access of Cyber-Physical Systems (CPS), this progress in comprehending complex activities using patterns can be further well realized and actuated, thereby allowing an interactive exchange between the physical and software components of the solution.

Automatic Video Surveillance (AVS) is a sub-field of scene understanding. AVS systems are specifically employed to identify human activities that are apparently abnormal or pose to be a threat, in the backdrop of a particular event. For example, bending down before jumping may pose to be a threat (suicidal behaviour) at a rapid-transit platform, but not (warm-up exercise) at an athletics' arena. CCTVs are installed at specific positions and the captured video data is constantly monitored by a software. Two purposes are served with the help of CCTV. Firstly, warnings (person roaming outside an ATM before attempting robbery, using abusive language before indulging in an aggressive fight, etc.) that led to a abnormal situation in the past, are useful for training purposes. It's indeed an effective way of learning for the authority, such that they can take necessary measures to prevent *similar* events actually takes place in the future. Thus, the classification model can adapt itself through learning past instances. Secondly, when surveillance is done in real-time, the emergency services can be automatically alerted of a potential risk by alarms, SMS or direct phone calls from the software system. Such systems do not necessitate a human operator to be present $24 \times 7$ before a camera, and thus eliminates out chances of overlook due to human-errors and operator fatigue. Hence, it can be ensured that whatever be the hardware configuration, the back-end software in an AVS system must perform effectively towards both the event classification and the human-activity recognition in real time. To successfully achieve both of these tasks (event classification and the human-activity recognition) efficient extraction of all the information (spatial, temporal, audio and text) from the video and their suitable representation are of prime importance. Thus, efficient exploitation of multimodal information leads to the correct identification of body-languages and human activities in a surveillance video.

## 1.1 Mathematical Representation and Multimodality Concepts

A video can be represented as a sequence of frames $\{f_k\}$, where $k \in [1, N]$ denotes the order (or, time) of appearance. Here, $N$ is the total number of frames in the video. Thus, considering $x$ and $y$ to be the horizontal and vertical position of a pixel within a frame, any pixel in a video can be referenced through a 3-D coordinate $(x, y, k)$. Now, videos typically contains multimodal information and they can be exploited efficiently through one/more of four major channels. These are elaborated subsequently, and shown in Fig. 1.

- *Spatial*. (i.e. space) Also known as the visual channel, this characterizes all the intra-frame information, i.e. within a particular video-frame $f_k$ and at a particular time instance $k$. Such apparent and single-image information can be perceived through image-processing methodologies, that ranges from inspecting RGB intensity values to localization of a certain object-of-interest.
- *Temporal*. (i.e. time) This channel characterizes the inter-frame correlations between successive frames at running time instances. Conventional image processing techniques are not suitable for computing such correlations between a pair of frames. However, a special video processing concept such as optical-flow needs to be exploited to capture information from a sequence of frames. This channel proves to be helpful in tracking iso-intensity pixels belonging to the same object, with passage of time.

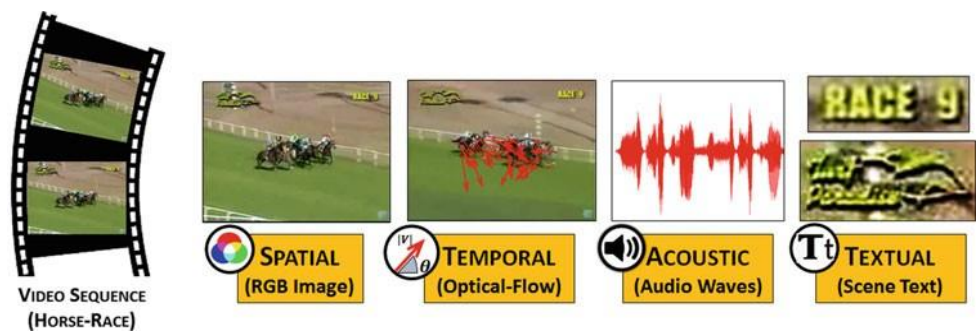

Fig. 1 The most common multimodal features, that are readily available from a video

- *Acoustic*. (i.e. sound) This channel refers to the auditory signals, that are heard while playing a video. The audio signal refers to background music (e.g. national anthem or birthday song), communication within two entities/groups (e.g. human chit-chatting, wolves howling), etc. Automatic speech recognition from an ongoing conversation can often prove to be highly useful in the context of scene understanding.
- *Textual*. In some cases acoustic conversation may not be recognizable as it is spoken in an unknown language or gibberish in nature. In such situations, text appearing on the video frame and/or subtitles can be extracted and recognized very efficiently by applying some classical image processing and classification techniques (Jana et al. 2017; Luo et al. 2019; Borisyuk et al. 2018). It may be noted that, video subtitles is fairly common among videos downloaded from social media and thus, it plays a crucial role in event recognition. Oftentimes, unstructured video description present alongside online videos, may also be useful in video classification (Kalra et al. 2019).

## 1.2 Advances over the Years

Over the years, the researchers have resorted to various strategies and classification networks to predict the ongoing activity in video data. Initial advances involved maneuvering patterns only in the visual or auditory channel, where spatial information in the frames were made use of to arrive at a decision. On the other hand, some of the approaches preferred essentially image-processing oriented methods such as Scale Invariant Feature Transform (SIFT) (Lowe 2004), or Histogram of Oriented Gradients (HOG) (Dalal & Triggs, 2005) to localize features. The fact that temporal features, such as the ones that accounted for long-term motion patterns in the video frames were needed to better characterize the semantic aspects of a video, was later realized. Various classification frameworks involving convolutional neural networks were used, with variant architectures and varying degrees of precision. It was observed that deep networks performed much better when they were given an information about the precise spatio-temporal location of an on-going action. By virtue of temporal action-localization, the 'essential' temporal segments were fed to a classifier network which correspondingly categorizes each such clip. Some further improvement was achieved, by means of fusing the spatial and temporal (and in some cases, even auditory) streams to make the classification more robust. Predominantly early fusion, late fusion, Borda-count and other approaches were used in this regard. This also played a role in clearly discriminating the closely-related classes of actions in the videos. In Sect. 2, we provide a detailed survey of all these salient approaches and fusion techniques that have been adapted for the task of effective event classification.

## 1.3 Motivation and Our Contributions

*Potential of Temporal Channel*. As put forward by Girdhar and Ramanan (2017), c/activity recognition in videos as a simple classification of constituent RGB frames, even if they proffer good results in select datasets. To perceive how an object behaves with time, it is not only essential to pinpoint the spatial location of the 'same' object in every frame it is visible in, but it is also beneficial if its periphery of locomotion (to which it is restricted, over the whole time-span of the video) can be determined. Thus in our work, firstly we illustrate how holistically encouraging results can be obtained by choosing to exploit only the temporal motion patterns in videos, using a CNN-RNN hybrid model. Interestingly, in this endeavor, the results are found to be at par with some of the spatio-temporal approaches in use. Next, we record the improvement in event recognition accuracy when spatial information is combined with temporal ones. Essentially, we try to validate the fact that the temporal features, if exploited suitably, can solely prove to be useful for classification. This is due to their enhanced persistence through motion across the frames of a video.

*Need for a Tailor-Made Fusion Strategy for Videos*. We also venture on a less-trodden path of trying to effectively fuse the deductions from the spatial and temporal wings, while retaining their individual merits. Traditional techniques of early-fusion, suffer from being unable to exert much influence on the classification accuracy with respect to the diversity of multi-stream convolutional neural networks (CNNs). To state it otherwise, when parallel CNN streams deal with diverse modalities (as for instance, RGB and acoustic data), one may negatively affect the others' training phase. On the other hand, late-fusion generally performs better (Lee et al. 2018) in combining disparate multimodal information. However, video as a whole, bears a close relationship with its constituent frames—the frames' event should incrementally make up the video's event, even if there are minute dissimilitudes. Not all late-fusion strategies are effective in integrating this frame-to-video relationship. In contrast to such methods, we perform a multi-tier fusion, where fusion is performed on both frame-level and video-level predictions, in an attempt to efficiently integrate them. Here, the concept of *conflation* (Hill 2011) is used to consolidate two discrete probability distributions into a single representative distribution at every significant level of the strategy. This proves to be extremely useful in cases where a consensus between the streams is desired. A bias-factor is also specified for this conflation technique. This plays a key role in promoting the most confident prediction which both the streams are in agreement with on pertinent levels, as opposed to the general practice of favoring *any* one stream. This takes care of fine-grained intra-class variations as well. A class which would be otherwise wrongly predicted herein has the potential to undergo the various stages of our multimodal fusion strategy to be properly recognized in the long run.

It is especially encouraging to see the approach surpassing other multimodal fusion methods in use, on benchmark datasets such as UCF-101 (Soomro et al. 2012), Human Motion Database (HMDB) (Kuehne et al. 2011), Columbia Consumer Video (CCV) (Jiang et al. 2011), and Kodak Consumer Video (KCV) (Loui et al. 2007). Overall, our approach proves to be minimally demanding regarding the quality of capturing devices/camcorders, eventually giving rise to an efficient Cyber-Physical System.

## 1.4 Organization of the Chapter

In this chapter, we delve deep into the intricacies that concern the task of video surveillance at hand, through event classification and human-activity recognition, and discuss its various aspects thorough the segments of this chapter. The remainder of this chapter is organized as follows. In Sect. 2, we provide a detailed view of the various approaches that have played a key role in shaping solutions to automated understanding of events/activities. These include early approaches aiming to uncover structured patterns from motion, advent of space-time descriptor based methods and contemporary state-of-the-art solution frameworks relying upon deep neural networks. The development of various architectures and how they contributed in efforts to address the shortcomings of their predecessors over time is extensively highlighted. In Sect. 3, we broadly portray our approach to robustly perform video scene identification with a greater deal of precision. Section 4 accounts for the implementation aspects and records the results of the experiments conducted on four major benchmark datasets for event classification. Our solution framework is seen to provide a greater rise in precision on each stage of the multi-level feature-fusion strategy as compared to the state-of-the-art methods in use. Finally, we conclude and discuss the future scope of work in Sect. 5, with a note on the enhanced scalability of our solution, for use in Cyber-Physical Systems and understanding of events in video surveillance.

## 2 Related Work

Many Cyber-Physical Systems use smart computing techniques and remain embedded in larger physical systems e.g., a car, an aeroplane or a building to perform specific tasks. Very often, this CPS interacts with the surrounding physical environment through use of actuators and sensors. These embedded CPS devices communicate with some cloud backbone through communication networks to help taking subsequent decision quickly and promptly. Use of convolutional neural networks for possibility of a fire breakout (Saeed et al. 2019) and application of machine learning and its variants such as ensemble learning and instant learning (Alzubi et al. 2018), has shown interesting results. Application of unsupervised learning using K-Nearest Neighbours for human age recognition (Priyadarshni et al. 2019) is also significant. Text and face recognition from videos captured in public places (Jain et al. 2020) and facial emotion detection (Mukhopadhyay et al. 2020) provide interesting work in predicting potential criminal activities. Automated vehicle localization through space-time descriptor models (Sharma et al. 2019) and time synchronization in vehicular networks (Ghosh et al. 2011) are important milestones in traffic surveillance. In modern era, the advent of Internet of Things (IoT) have further revolutionized lives (Tanwar 2020). Several applications of IoT in different domains like, pharmaceutical fraternity (Singh et al. 2020) and healthcare in elderly patients (Padikkapparambil et al. 2020; Singh et al. 2020) have shown positive social impact. Bringing innovation to homes and common life, efficacy of IoT in household waste management (Dubey et al. 2020) and reliability analysis of wireless links for applications under shadow fading conditions (Sehgal et al. 2020) are worth mentioning. It thus may be pictured that the profoundly intertwined domains of computer vision and machine learning find heavy manifestation in fields of CPS and IoT, thereby changing everyday lives by leaps and bounds. With technological advances in a more demanding society, cyber crimes and threats of compromised security and privacy have become a concern too. The widespread popularity of hand-held devices including smartphones, laptops, palmtops is providing immense benefit of ubiquitous computing on the fly. However, this mass-scale computing also suffer from many cybersecurity challenges (Jana & Bandyopadhyay, 2013, 2015; Gupta et al., 2020) including phishing, IP masquerade attacks, denial-of-service attacks, mobile ad-hoc network security threats. Information security incurs major challenge when Cyber-Physical Systems avail cloud-based service, distributed throughout public domain of Internet (Ukil et al. 2013). Any Cyber-Physical System dealing with Inter-net of Things (IoT), exchange huge number of messages over Internet that requires a high-volume distributed file system to act seamlessly with reliable fault-tolerance (Paul et al. 2016). With these premises, we now proceed towards work on event and human-activity recognition in computer vision applied in video surveillance domain.

The problem of event recognition has garnered widespread interest in research in computer vision. The event recognition datasets that we work on, are replete with content diversity encompassing classes pertaining to domains. Examples include humanitarian basic action, outdoor scenes and indoor activities that may even occasionally feature various animals in action. This motivates us to provide a backdrop for discussion of the various approaches that researches have resorted to over the years. In subsequent sections, we highlight many of these significant approaches across time.

## 2.1 Early Concepts of "Structure from Motion"

One of the earliest and most intriguing works were by Potter (1976) in the year 1976, which being a paper in transcendental psychology, presented assessments of the human mind's ability to retain information about a series of pictures flashed in a rapid sequence. Evidently, the *gist of a video*, or what may be called a brief understanding of complex content was being studied. Speculations are that, this planted the seeds of interest in the then newly conceived discipline of computer vision, whereupon the aim was to automate the understanding of these events or in essence, high-level complex activities. In the early 1990s, methods were being developed to tackle the very basics, the smallest of gestures, for instance "the raising or moving of a hand". Analysis of the body posture featured as a recurrent theme in many of these works. Psychological studies once again lit the scenario up, as the works of Johansson showed that gait analysis and motion could be tracked through Moving Light Displays (MLD), that were not in the least affected by external distractive features such as the lack of illumination in the video scene, or jitter, semantic spatial background variation. For instance, an MLD would specifically not distinguish between a man jumping in a forest and a man jumping on an asphalt road, since the motion is all that bears emphasis for an MLD system, which is in this case of course, jumping. Some of the contemporary approaches to tackle the problem in the field of computer vision, wound around the aim of automated interpretation of MLD results. As discussed in the works of Cedras and Shah (1995), quite a few of the works were found to be based on the concept of efforts to quantify Structure From Motion (SFM). SFM relied on exploiting the three-dimensional coordinates of all the objects in motion in the input video that would reveal some information about the way they execute their motion. One of the frontiers that accelerated the progress in this regime was *optical-flow*. A method that made computational sense of the pattern of motion of the objects, depending on the way their movement changed across successive frames in a video, optical-flow has stood the test of time, still a go-to resort for many enthusiasts in the field. The fact that it also accounts for the change of the object movements with sustainable camera movement, makes it an effective SFM method for articulating motion over time. Mathematically, an optical-flow vector could be interpreted as a two-dimensional vector field with each entry signifying the displacement of the particular point in consideration from one frame to the successive one. The method is powerful in the sense that it provides considerable information about the velocity of movement, and the part of the frame featuring the said object. As an immediate corollary, it found its practice and use in the domain of surveillance and effective tracking of vehicular motion. This paved the way for many more advances in the allied fields of object tracking. Many standard methods are in vogue for computation of optical-flow, some of the most prominent being the works of Horn and Schunck (1993), and Oron et al. (2014), being predated by the traditional methods of block-based and phase-correlation methods.

A number of methods followed the advent of the optical-flow algorithm, which were trajectory-based. This served as a ground for computational realization of relative motion of objects in a single video frame. It served as a citation of the potential power of temporal features, which until then had often either been a moot-point or viewed as a mere auxiliary of the spatial information a video had to offer. Accuracy is at times, seen to fluctuate, for methods based on optical-flow are prone to be affected by low illumination, cluttered background and instances of occlusion present in the videos. Clarity is sought primarily in a number of approaches, where ambiguity is to be eradicated in the discovered trajectory points. Even then, it serves as one of the most prominent means of accounting for the temporal features that have a telling effect on recognizing the activity in the video. At this juncture, Elgammal et al. (2000) suggested segregation of images into two salient parts, namely the background and the foreground, could turn out to be useful in tackling the problem of recognizing events. To reinforce the conceived idea, instances of identifying the human body structure from silhouette-images or frames. Thus, background subtraction to focus on the entity in the image executing motion was of interest. The silhouettes in turn, can be uncovered using now renowned algorithms such as the Grab Cut algorithm proposed by Rother et al. (2004) that can separate the foreground object from the ambient background. The contour of the foreground object involved in the activity can thus be analyzed in this approach, thereby helping one to assess the impact of considering the size, shape and pattern to realize the. The extracted silhouettes can serve as useful information, emphatically so when human activities are involved, so that the gait patterns may be studied. It was at this time that a pioneering work by Polana and Nelson (1994) dealt with non-parametric approaches for machine perception of activities. Before moving on to the discussion of the various three-dimensional, local-descriptor based methods to quantify actions, it is worth mentioning the two-dimensional formulations the authors came up with. Motion was first articulated through temporal difference images, and Human presence was detected through the study of both either extracted silhouettes, as well as from a *trajectory primal sketch* obtained from the MLD applications, inspired by the works of Gould and Shah (1989). "Action" was an entity that they classified into two main kinds—*stationary* and *non-stationary*, owing to the fact that stationary events such as sitting would not cause significant geometric translation of the person executing the motion, as compared to non-stationary ones such as walking, running, playing and so on. Actions of different kinds relevant to the dataset are marked with varying levels of periodicity by the nature in which they are executed, whereupon periodicity in the action video is detected using Fourier Theory. According to the concept, actions found to be considerably periodic are detected by this approach. An impressive account which builds on this concept is presented by Bobick and Davis (2001), in which the activities and the fine transitions between them are given increasing importance, and first individually mapped onto a representation the users call a *temporal template*. The temporal templates across the length of frames are then aggregated effectively to express the action, in its making. On giving equal weight to each of these constituent frames, we are led to the representation of the *Motion Energy Image* (MEI) template. While a MEI can traditionally stand for the intensity of the activity, on the other hand another informative portrayal is found in the *Motion History Image* (MHI) where one can get an idea to how exactly the particular motion was executed over time. As an example, it could map the sequential forearm movements of a tennis player distinctly depending on the kind of shots he/she chooses to play. These templates prove to be one of the most prominent amongst the early advances of tackling the problem of event and action understanding significantly in video data, and form the basis for further enhanced studies on the same.

## 2.2 Space–Time Descriptors

The progress of understanding high level activities in video content, took a rise with proper interpretations of *features*, which could now venture along space and time fields. The *Harris Corner Detector* in its two-dimensional form, was used to demarcate spatial points in an image, with a degree of content intensity variation above a certain threshold, as discussed in the work by Derpanis (2004). A powerful extension was provided in the renowned work by Laptev (2005), in which a three-dimensional extension of the basic Harris Corner Detector was proposed. The inclusion of a third dimension accounted for the fact that now temporal relevance would be highly recognized as a factor for recording regions of change in the frame image. A fresh concept of "Space-Time" or "Spatio-Temporal" Interest Points (STIP) was obtained through this perspective, and these interest points were profoundly insightful to exploit since they carried the spatial content information, as well as ensured the temporal relevance of the motion pattern that sustained the content change. A beautiful illustration that could help us better comprehend the phenomenon could be through an example, as in Fig. 2.

The intriguing fact worth mentioning in this approach that the points where temporal movement is considered noteworthy, across the length of the frames for the video, instead of just accounting for spatial changes such as appearance of a new object in the scene (which in this case, may be thought to be the ball). The first frame where the ball enters the scene, causing content change does not locate it as a potential spatio-temporal interest point just then, because it does not contribute any change to the location of the region in which it appears in the image. Across the frames, the batsman does not let his head stance position vary too much, which evades it from being a strong point of interest, and gets shadowed by other points in its vicinity. These points evidently undergo a comparatively effective displacement in the dimensions with respect to the boundary points, in the extended formulation of the Harris Corner Detector. These are the points which execute motion significant enough to portray the characteristics of the action class (in this case, Cricket Shot). Thus, although the portion of the image containing the entrant ball is not initially labeled "a point of interest", it gets included in afterwards, when the ball is hit by the batsman, explicitly indicating the motion of the cricket shot being played.

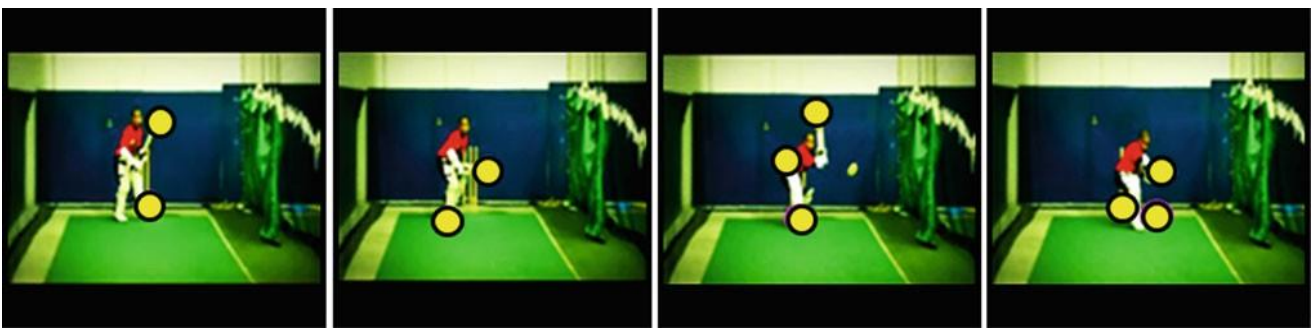

Fig. 2 An illustration of how *Spatio-Temporal Interest Points* (marked in yellow circles) obtained through the extended 3D Harris Corner Detector could be key to comprehending actions manifested across frames across video (in selection: 'Cricket Shot' from UCF-101 Soomro et al. (2012) dataset)

Based on this extension, Blank et al. (2005) deviseda novel strategy to not only mark the points of action, but also to model the activity in transit. For instance, the motivation here was to discover what kind of a spatio-temporal volume could imply a particular action class. At this juncture, one might rightfully draw a few parallels with the work of Polana and Nelson (1994), whereupon the authors interpreted actions to be periodic. The merit here, with mathematically articulating the way an action evolves over time, one is able to eradicate the over-dependence on periodicity, and is able to recognize it instant whenever it is a trait of the particular activity class. Limb movements in walking, or repeated structure of motion patterns as seen in cycling, might serve as particularly relevant examples. Once visualized geometrically, the scenario translates into the idea of a spatio-temporal cuboid. This maps the degree of movement seen in the video through representative blocks sampled over parameters (space, time and volume scales). In a work by Wang et al. (2009), the efficacy of these important points-of-interest are elevated by trying to extrapolate the Hessian detector to a third dimensional strategy, in a manner akin to the Harris detector. It is seen in hindsight, that these cuboids, serving as local descriptors may be used to map some motion trajectories from which motion that exerts dominating effect on classification of its kind or scene understanding, can be uncovered. However with unconstrained videos, shot with lack of proper professionalism, there may be occasionally jittery content, and the dexterity of the local descriptor approach is found to be debatable in such cases. There might be relative motion in the actions, and there might be camera-movement in the video sequences as well, leading to motion depth considerations studied in considerable detail by Jiang et al. (2012). This also paves the way for a discussion of the more recent approaches to the problem of event recognition in unconstrained video data, many of which make use of deep neural network architectures in various forms.

## 2.3 Recent Approaches Using Deep Neural Architectures

Having adopted many strategies based on mapped templates, hand-crafted features, and local descriptors over a span of a decade or more, the research community felt the urge to try using deep learning models for a better understanding of the unconstrained video scene. Convolutional Neural Networks (CNNs) proved to be stunning in their work of classifying images, as various architectures of CNNs continued to ace the Imagenet Classification Challenge Russakovsky et al. (2015), each subsequent neural network model appearing to be more proficient in the task of classification than its predecessor. This excellence of CNNs in general while dealing with image content, served as further motivation to deploy them in the task of video recognition. Another inclination which contributed to the rise of deep learning models in the solution of this problem was the fact that hand-crafted features, when learnt, were often found to be reliant on the video quality in general. These also fell short in accounting for the widespread content diversity of classes that the activities or events occurring in the videos. This was observed to be true in the case even where these individual features were aggregated through a dictionary-based codebook or vocabulary of visual features, that attempted to extensively markdown some distinguishing action patterns for each of the applicable classes. The approach was named the *Bag of Visual Words* (BoV), and efforts to strengthen it were made by making the features being aggregated powerful, with techniques such as *Histogram of Gradients* (HoG) and *Vector of Locally Aggregated Descriptors* (VLAD) being used. Although temporal relevance was given some attention, in a modified approach, it called for more significant importance, due to the sheer amount of enormous possibilities and diverse semantic characterization that each of the activity classes could potentially bear in a dataset.

Some of the initial CNN networks, intuitively exploited the spatial features that the videos contained, as the models were fed with successive frames of the video. Alternatively, this was also effectuated by performing image classification and/or semantic segmentation Mukherjee et al. (2020) on the individual frames. Such spatial techniques can correlate a scene's subject to the background, thereby realizing the scene content; as for instance, the common appearances of a soccer ball in a stadium or gallery, and the fact that a football match could be prevalent in the event. Intriguingly, this instance could lead one to dawn upon the fact that the spatial features alone could often be misleading for the task of classifying scenes. An intuitive example, would be the fact that the spatial information "stadium" or "green grassy fields" could relate to more than just one class of events, such as "soccer", "cricket", "baseball", or for that matter any generic outdoor sport activity. So, spatial CNNs were in vogue across various architectures, until the growing importance of temporal features were something that could no longer be avoided. Temporal features would directly correlate with the limb movements and other respective motion patterns, associated with the particular action. There could be ways of exploiting this through SFM methods that we discuss in the earlier section of approaches. The processed image data obtained as a result of the applied SFM technique can be then fed as input to the CNN model. In their work, Zhang and Xiang (2020) make use of SIFT, HOG and *local binary patterns* (lbp) to localize the temporal information, that is the persons in the image executing the action. A temporal CNN could be designed across a variety of CNN architectures. Some authors while studying temporal features resort to Recurrent Neural Network (RNN) architectures. As the name suggests, they are capable of accurately picking up long term motion implications and specific pattern sequences that correlate to some classes. The fact that these recurrent architectures have various memory cell units integrated that help them to retain sustainable information, and a feedback layer which ensures the learning is robust across time samples to account for the entire length of the data. Approaches relying exclusively on temporal features for video classification are substantially less, and we (Jana et al. 2019) in one of our works, demonstrate the immense potential they have as sole performers itself, till an extent where we show them to be not only capable of outperforming standard exclusively temporal methods, but also being at par with some of the spatio-temporal techniques of classification. This is because of the hybrid deep learning model that we use. Choosing a set of representative key-frames that represent the video content for each class with minimized temporal redundancy, we perform optical-flow on these selected key-frames. To localize the robustly significant actions that characterize the activities, we feed the magnitude images of these optical-flow images to our deep learning framework, which comprises of a CNN-RNN (LSTM) fusion. This signifies the fact that the output predictions of the CNN model serves as input vector, duly reshaped for the RNN framework. At this juncture, it is worth mentioning that we use the ResNet-50 architecture as the CNN, which being a residual neural network exploits action features considerably from the input magnitude images, being considerably "deeper" than the likes of VGG and Inception-Net that shortly precede the advent of strong residual blocks in a deep neural network. This speaks for the fact that a CNN-LSTM fusion for temporal features proves to be more than dexterous in the task of classifying videos into action classes.

However intensely dedicated these networks might be for the task of recognizing exclusively spatial or only temporal features from a video in order to furnish optimized classification results, it was realized that the two indeed had to go hand in hand. Thus rose the concept of multimodal deep fusion techniques, that contributed to a lot of interesting studies and some really notable advance in terms of identifying action classes from disperse datasets. Various methods of multimodal fusion are some of the foremost approaches that yield truly encouraging results in comparison to the standard uni-stream ones. The authors exploit parameters across three main streams (visual, action and auditory stream) in an intriguing approach of "Who", "What", and "Where", in an effort to characterize the scene content (Zhang & Xiang, 2020). The deep learning framework they come up with is a multi-stream CNN with features being globally pooled across the respective frameworks, to give multimodal cues.

Figure 3 articulately demonstrates the need for a "spatio-temporal" fusion approach in solving or designing a deep neural network in order to learn the characteristics of the various activity classes in the datasets. It is interesting to see how the frames of videos essentially belonging to centrally similar theme (kicking/soccer) but evidently they are distinct classes, as it can be seen. This speaks for itself why exclusive spatial information such as a lush green field, stadium, or even sole temporal features such as foreword limb movement for the task of kicking are not sufficient to accurately classify the event since "Soccer Juggling" would have minutely different motion intensity pattern than that of a "Kicking" class, which might be done in any kind of an ambiance and not necessarily on a field. 3-D Convolutional Neural Networks are statistically seen to outperform their two-dimensional counterparts, due to three dimensional kernels, which helps extract features corresponding to a third axis, denoting time. A host of approaches present score and rank-level fusion, often on accuracy values obtained on separate streams of the multimodal network, but this quite naturally leaves behind a semantic gap to be bridged. On the other hand, some of the approaches by Luo et al. (2019), and Li et al. (2020) attempt to exploit the spatio-temporal features through fusion of the respective softmax scores of the deep learning model. Another set of salient approaches perform weighted conglomeration of the extracted features from each of the representative multimodal streams, be it visual, motion or auditory, in a secondary max-pooling layer of the network, with an intention of thereby promoting the best of those extracted across each wing, as is the motivation seen in Cherian and Gould (2019). The reason these models sometimes fail to offer the precise classification result for an event class, is due to the underlying ambiguity of the *degree of importance* that features uncovered from a particular stream or wing deserves, for instance the auditory features extracted in a particular case might be promoted largely due to its intensity despite the fact that it is misleading and could have might as well be suppressed.

An interesting study of the concept of *coherence*, that gives an assertive suggestion on when one may opt to suppress the temporal stream and focus on the exploitation of the spatial wing is due to the Siamese Network, conceived by Lu et al. (2017). As shown parallely in the works of Varior et al. (2016), Siamese Networks have the ability to take entire video sequences as input and decide on whether the temporal variation in the frames is significant enough to be given an enhanced priority over the visual stream, or vice-versa. To eradicate such possible anomalies comprehensively, during fusion we (Jana et al. 2019) resort to a multi-tier conflation based strategy that provides us with a noteworthy increase in the number of correct classifications, as we achieve the con-

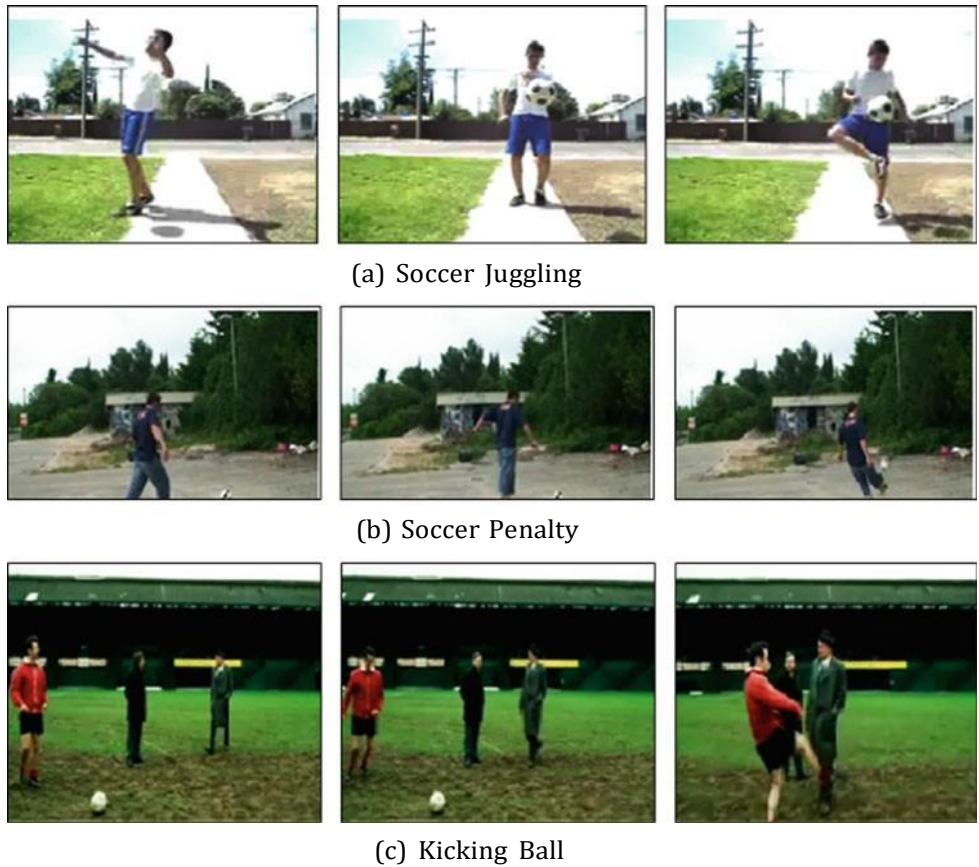

(a) Soccer Juggling

(b) Soccer Penalty

(c) Kicking Ball

Fig. 3 Illustrating the need of a *spatio-temporal fusion network* rather than an exclusive one, to accurately identify event classes, related to a central theme (in this case, kicking/soccer)

sensus from both video and frame levels. Closely related to the concept of spatio-temporal fusion, an illuminating work on the domain by Karpathy et al. (2014), while working with the UCF-Sports dataset, deals with a concept of fusion coupled with a foveated architecture. Apart from spatio-temporal streams, the deep learning CNN model is equipped with a "*fovea*" stream, which stands for *central view*. The fovea stream contains a central crop of the frame image of the input video, somewhat akin to a *zoom-in* effect. It plays the role of substantial auxiliary information that guides the CNN with some added details that prove crucial to characterize the event, as we show in Fig. 4. This was seen to improve recognition results. However it is quite intuitive that the interpretation of the foveated image as a centrally cropped image from the main frame is better suitable for sports classes in general, than for generic event identification. The puzzle of having a competent enough fovea stream for a broader spectrum of events is a problem that is being currently worked upon, as the concept of an added "attention" to what is usually at offer across the multimodal streams, is believed to be able to boost the classification results by a considerable margin. This can undoubtedly have a telling effect on many areas of Cyber-

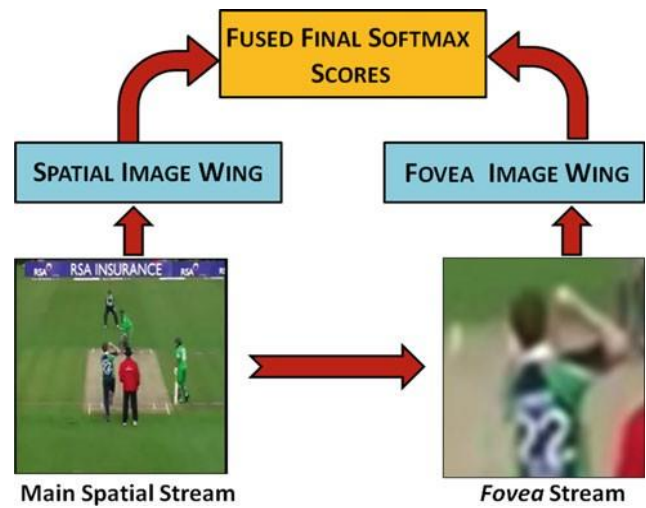

Fig. 4 Deep architecture with *fovea* stream by Karpathy et al. (2014)

Physical Systems and surveillance, since it helps in pinpointing the action of important interest, and thereby tracking it with a greater degree of precision.

## 3 Video Scene Identification

We start our event-recognition pipeline with reduction of the space of frames. This is done by a systematized key-frame selection technique (Sect. 3.1). Firstly temporal redundancy is reduced by eliminating redundant information. Then the most distinctive set of frames are chosen by a graph-based approach. These form a representative set of frames for the video. This is followed by extracting multimodal information from those key-frames (Sect. 3.2). Next, these multimodal features are maneuvered through separate dedicated deep architectures. Finally, the respective frame-level and video-level predictions are fused (Sect. 3.3) from each such modality channel, to proffer a final decision pertaining to a video.

### 3.1 Key-Frame Sampling

In all videos, the physical setting or the backdrop remains constant throughout majority of the duration. But, there are specific time-spans of the video when the scene changes, either gently (*gradual transition*) or in suddenness (*abrupt transition*). As such, the whole time-duration of a video can be broken down into a number of semantically-meaningful fundamental structural units, known as *shots* (Mohanta et al. 2011). These can be regarded as a set of frames, that appear at successive time-stamps and have similar spatial appearance, when the video is played. Now to obtain a *storyboard* from a video, it is essential to retrieve one/more representative frame(s) from each such shot. These are collectively known as *key-frames*. The better the set of key-frames chosen, more is their viability to be regarded as a summary of the whole video.

Upon more deliberate thoughts on the underlying concept behind representative frames, another inherent property becomes apparent. The number of key-frames chosen from a fixed time-span should be directly proportional to the number of key-frames chosen from that span. As for example, let's suppose that the frames of a sixty second video practically shows no considerable motion during the first fifty seconds, and all the actions are packed in the last ten seconds. In this case, most of the key-frames should ideally be chosen from the last ten seconds (even though its span is smaller), and a handful from the first fifty seconds. We follow a two-stage process (Jana et al. 2019) to sample a set of $n_{KF}$ key-frames from a video, as explained in the subsequent paragraphs.

***Reduction of Temporal Redundancy***. As we understand, frame-sampling should be from the temporal (time) axis. Firstly, we reduce the temporal redundancy by replacing the *similar* (with respect to motion of constituent pixels) frames, by a single representative frame. This is started off by computing the dense optical-flow (by any classical algorithm such as, Horn–Schunck (1993) or Lucas–Kanade (2014) between consecutive pair of frames. Thereby with each pixel $(x, y, k)$, a flow-vector $(u\delta k, v\delta k)$ is associated that corresponds to its spatio-temporal displacement to a new location, $(x + u\delta k, y + v\delta k, k + \delta k)$. For consecutive frames, $\delta k = 1$ and thus, we can represent the magnitude and slope of flow-vector as, $\sqrt{u^2 + v^2}$ and $\tan^{-1}(v/u)$ respectively. Further for a frame, the overall distribution of motion can be quantified by histograms of magnitudes and slopes of corresponding flow-vectors of constituent pixels. These two histograms are then concatenated, to represent a frame's motion (henceforth called as *motion-histogram*, by us). The next task is to determine if consecutive frames are homogeneous with respect to motion. If that is the case, we may not take both but discard one of them. We, therefore, record the temporal disparity ($td_{k \to k+1}$) between successive pair of frames, $I(:, :, k)$ and $I(:, :, k + 1)$, as the $l_1$-norm of corresponding motion-histograms. When this temporal disparity ($td_{k \to k+1}$) is less than a certain threshold, it can be adjudged that the corresponding frames are *temporally redundant*—thus, one of them is discarded. An optimal value of the minimum threshold ($td_{min}$) is statistically found to be

$$td_{min} = \overline{td} - \hat{\sigma}_{td} \quad (1)$$

where, $\overline{td} = \frac{1}{N-1}\sum_{i=1}^{N-1} td_{i \to i+1}$ and $\hat{\sigma}_{td}^2 = \frac{1}{(N-1)-1}$

$\sum_{i=1}^{N-1} (td_{i \to i+1} - \overline{td})^2$. At the end of this step, we get a subset of frames from the video that are temporally distinct w.r.t. associated motion of pixels. The search-space of key-frames is thus drastically reduced, making our subsequent step computationally efficient and effective.

***Selection of Distinctive Frames***. This stage begins with the subset ($S$) comprising of temporally distinct frames, from the previous step. Firstly, a *complete graph* is formed each of whose vertex corresponds to a frame from $S$. For any two vertices (frames) $v_i$ and $v_j$, the weight of the edge connecting them is the temporal disparity $td_{i \to j}$, i.e. the $l_1$-norm of corresponding motion-histograms. Thus, the edge-weight can be physically interpreted as the temporal (w.r.t. motion of constituent pixels) distinctiveness between frames—edge-weight increases with disparities in flow-vectors.

But, maximizing distinctiveness amongst frames is not sufficient for an effective key-frame selection algorithm. Let us consider the example yet again, that we gave in the introductory paragraphs of this section—a 60 s video, with no considerable motion during the first 50 s, and all the actions packed throughout the last 10 s. Since all our frames in the last 10 s change most rapidly, an algorithm solely reliant upon temporal distinctiveness would be biased to choose frames that are in close time-proximity to one another (within the last 10 s).

Thus, our motive of story-boarding the whole video will not be satisfied here if we rely only upon temporal distinctiveness. To solve this problem, we store the timestamp of each frame corresponding to the node it represents in the complete graph. And in each iteration while choosing the most distant (distinctive) pair of nodes (frames), we pay a close attention to these timestamp values to decide upon whether to include the edge they represent, or not. This condition is regarded as the '*viability*' of the edge. An edge can be chosen in a particular iteration if and only if it is viable, and moreover, edges once marked as non-viable are not considered for inclusion in any of the further iterations. For an edge to be considered as *viable* for inclusion, two conditions must be satisfied:

- Difference in timestamps between terminal nodes of the edge should be above $d_{low}$.
- Difference in timestamps between each of the terminal nodes of the edge and any previously chosen node, should be above $d_{low}$.

Here, $d_{low}$ is the minimum acceptable time-gap between any two chosen key-frames and its value is $\left\lceil \frac{N}{2 \times n_{KF} - 1} \right\rceil$. As we defined in our introductory paragraphs, $n_{KF}$ is the preset and fixed number of key-frames, we want to represent a video by. This value arises after allowing a relaxation on the stringent time-gap of $\left\lceil \frac{N}{n_{KF} - 1} \right\rceil$ that must be satisfied when we want the frames to be equally spaced across the time-axis.

In this way, frames are selected from the graph until $\frac{n_{KF}}{2}$ iterations are completed, or all edges are marked as un-'viable', whichever occurs first. In each iteration, one edge i.e. two vertices are selected.

This stage is explained pictorially in Fig. 5. We consider a video consisting of 2195 frames, in total. Supposing that only ten frames were obtained in subset (*S*) (comprising of temporally distinct frames) from the first stage, a complete graph of ten vertices were formed initially. Also, considering that we require $n_{KF} = 6$ key-frames to be chosen from the video, there would be a maximum of $\frac{n_{KF}}{2} = 3$ iterations. Thereby, in this case, the minimum acceptable time-gap ($d_{low}$) between any two chosen key-frames takes the value of $\left\lceil \frac{N}{2 \times n_{KF} - 1} \right\rceil = \left\lceil \frac{2195}{2 \times 6 - 1} \right\rceil = 200$. For each iteration, all the 'viable' edges are shown as *black solid lines* in Fig. 5. Further, the most distant pair of vertices from amongst the viable edges are shown by a *superimposed red dotted line*. Terminal vertices of already chosen edges are shown to be *ticked*. At the end of three iterations, the chosen edges are (0, 5), (4, 7) and (2, 3), in order. The six key-frames chosen through the three iterations, are displayed again at the bottom with their respective timestamps. It is evident that the time-span $t = 455$ to $t = 1443$ is more action-packed than $t = 1444$ to $t = 2195$, because more key-frames were chosen from the former.

## 3.2 Realization of Spatio-Temporal Features

Two types of features are focussed on, in the current work viz., *spatial* and *temporal*. Moreover, we are only interested in the features extracted from the key-frames. The spatial feature is associated with the space of a single frame. It is realized by the raw RGB-image form of a key-frame. On the other hand, the temporal feature is associated with the transition from one frame to another. To retrieve the temporal feature from a frame $I(:,:,k)$, the dense optical-flow (by any classical algorithm such as, Horn–Schunck (1993) or Lucas–Kanade (2014) is calculated between $I(:,:,k)$ and its immediate next frame, $I(:,:,k+1)$. The temporal feature is represented by the grayscale image where each pixel position $(x, y)$ gets the magnitude of the flow-vectors at $I(x, y, k \rightarrow k + 1)$. The magnitude value is normalized in the range [0, 255].

## 3.3 Deep Neural Architecture and Decision-Fusion

After extracting multimodal information (here, spatial and temporal), there is a need to combine these such as to get a single probability distribution (for logistic regression) or, a single prediction class pertaining to a video. Keeping deep learning based classification approaches in mind, broadly, the fusion (Liu et al. 2018) strategies can be classified under the following subheadings:

- **Feature-Level Fusion**. This is also known as *early fusion* or *data fusion*. It creates a combined feature-vector representation by *concatenating* all the input feature vectors, corresponding to each of the modalities. Subsequently, this combined feature-vector is trained on a single classifier model. The classifier model must be compatible to tackle all the modalities simultaneously. Moreover, this necessitates the combined vector to be pre-processed separately and aptly, such that it is suitable for the single classifier model.
- **Kernel-Level Fusion**. Here, the fusion operation is closely interleaved with the training of classifier. Kernel function is a mapping, that is applied on a set of non-linearly-separable data-points to map them to a higher dimensional space (such that now, they gets linearly separable by a hyperplane). As such, they enables a linear classifier to be applicable towards linearly-inseparable data-points. In this fusion strategy, corresponding to the feature sets from each modality, the kernel-values are calculated separately each time. After this, they are integrated to form a new composite kernel-function.
- **Decision-Level Fusion**. This is also known as *late fusion*. It combines the unimodal predictions (class label/probability distribution) from the last fully-connected layer of each single-frame classifier stream (each relied upon for different modalities) of an ensemble network. Since this is used to consolidate scores *post-classification*, researchers (Rana 2011; Pinar et al. 2016) are of the opinion that the raw-data level correlations are not properly exploited. Nevertheless, this allows to have a separate classifier model, each well-suited to tackle each modality.

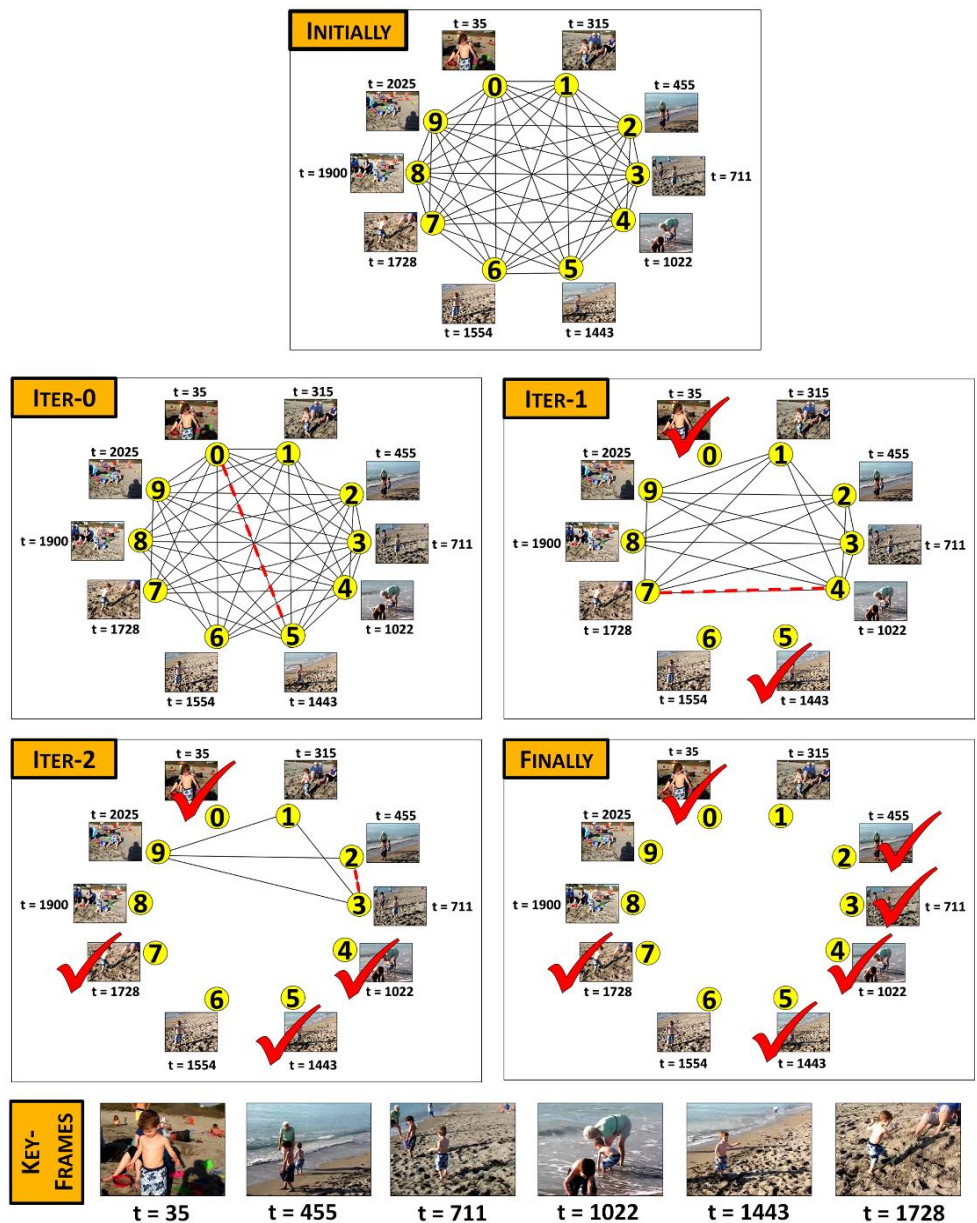

Fig. 5 Demonstration of the stage of "Distinctive Frame Selection" subsequent to reduction of temporal redundancy. The chosen key-frames are displayed at the bottom with their respective timestamps of appearance

Regarding the deep neural architecture, we employ a hybrid architecture (Jana et al. 2019), comprising of a Convolutional Neural Network (CNN) and Recurrent Neural Network (RNN). While the CNN portion is manifested by ResNet50 model, we use a LSTM network to realize the RNN counterpart. The flow of data through this hybrid model goes like this: (i) A certain modality information (spatial or temporal) is exploited from each of the key-frames pertaining to a video, and fed to the CNN (ii) Corresponding to that particular modality (spatial or temporal) for each key-frame, the CNN provides us with a prediction vector at the last fully-connected layer and one/more feature vectors from the preceding fully-connected layer(s). The prediction is in the form of a $L$-class probability distribution, where the $l$th value represents the probability of that frame to belong to the $l$th category/class. These are henceforth regarded as *frame-level prediction*s. (iii) Next, the frame-level features of all the key-frames pertaining to a video are accumulated. They are reshaped appropriately to three-dimension before being fed to an LSTM network. (iv)

The LSTM, in turn, outputs a single probability distribution for a video that ideally acknowledges prediction of each of its constituent key-frame. This prediction is also in the form of a $L$-class probability distribution. But in contrast to the frame-wise prediction, the $l$th value represents the probability of the whole video to belong to the $l$th category/class. Thus, we refer this as a *video-level prediction*.

Next concern is to consolidate all the frame-level probability distributions of constituent key-frames and the video-level prediction, for each modality. Most researchers (Wu et al., 2015; Peng et al., 2017; Cherian & Gould, 2019) prefer to use average of probability distributions, to integrate them. But, it does not always provide a desirable solution. This is because the consolidated distribution can give a completely new distribution, that is consistent with none of the initial distributions. So, we prefer to use a *biased conflation* (Jana et al. 2019) of probability distributions. Moreover, making the whole process multi-tier gives us a systematic approach that integrates frame-level and video-level predictions effectively.
*Biased-Conflation.* For two probability distributions $P_1$ and $P_2$, their conflation is defined as,

$$P_{\text{conflated}}(X = a) = \frac{P_1(X = a) \times P_2(X = a)}{\sum_{b=1}^{X_{max}} P_1(X = b) \times P_2(X = b)} \quad (2)$$

To this, we adopt a biasing technique to make it close to reality (i.e., one of the initial distributions). The Bhattacharyya distance (Bhattacharyya 1946) from $P_{\text{conflated}}$ to $P_1$ and from $P_{\text{conflated}}$ to $P_2$ are computed. Thereafter, the conflated distribution from $P_{\text{conflated}}$ is biased to the closer distribution (from amongst $P_1$ and $P_2$) with an appropriate biasing factor, dependent on these two distances.
*Cross-Fusion and Self-Fusion.* Apart from the concept of biased-conflation, we introduce two types of fusion. *Cross-Fusion* is defined as the biased-conflation of probability distributions, belonging to two different modalities. *Self-Fusion* is defined as the biased-conflation of the probability distributions, corresponding to all the key-frames of a video. We go on aggregating distributions in a hierarchical and multi-tier scheme (Jana et al. 2019). Effectively, cross-fusion of frame-level (or, video-level) distributions give similar frame-level (or, video-level) distribution. On the contrary, self-fusion of frame-level distributions return video-level distributions. So, cross-fusions are applied in the same hierarchy, while self-fusion is applied to move up the hierarchy ladder.

## 4 Experimental Results and Discussion

As we stated previously, an efficient Cyber-Physical System (CPS) intended towards video surveillance, depends heavily on the capability of its backend algorithm. To state it otherwise, this algorithm should be equally proficient towards scene understanding or event classification on one hand, and human activity recognition, on the other. Moreover, since video capturing devices are often inexpensive, the recorded videos suffer from serious quality issues. Thus to acknowledge how the proposed method fares towards these surveillance systems, we evaluate our results on the following datasets:

(i) **Columbia Consumer Videos (CCV)** (Jiang et al. 2011): This dataset is consisted of 9317 unedited consumer videos from YouTube, that can be sub-classified into 20 event categories. In this collection, low-level events like, locomotive objects ("bird", "cat") and outdoor scenes ("beach", "playground") coexist side-by-side with high-level events like, sports ("biking", "ice-skating"), and social gatherings ("parade, "birthday").
(ii) **Kodak Consumer Videos (KCV)** (Loui et al. 2007): This dataset includes 1358 quality-degraded videos from the actual consumers of Eastman Kodak Company's products, and 1873 videos from YouTube. These videos are spread over the broad categories of activities, occasions, scene, object, people and sound, that accounts for a total of 29 event-concepts.
(iii) **UCF-101** (Soomro et al. 2012): This dataset includes 13,320 short clips from YouTube distributed across 101 action classes. Although the videos does not go beyond realistic human-action, the diversity in actions ranging from small-scale facial movements to large-scale locomotory activities, is what makes this dataset challenging.
(iv) **Human Motion Database (HMDB-51)** (Kuehne et al. 2011): This has 7000 clips spread over 51 human-action categories, collected from numerous freely-available movies on Prelinger archive, Google and YouTube. Videos showcase quality aspects, characteristic variance in duration and have stabilization issues.

While the first two datasets deal with unconstrained videos and event classification, the latter two are focused on human-activity recognition. Thereby, we aim to provide a exhaustive performance evaluation of our method.

We have based our evaluation in two stages. We told in our motivation (Sect. 1.3) that we intend to show the potential of a standalone temporal stream. The first stage evaluation is based upon this idea. With this in mind, in Table 1, we tabulate some of the past and recent state-of-the-art (SoTA) methods that focussed exclusively on temporal feature. With these methods, we compare the performance of our temporal CNN-LSTM stream. Although our method surpasses all the tabulated SoTA methods in event recognition on CCV and KCV datasets, there are still scopes of improvement in the temporal-based human-activity classification on UCF-101 and HMDB-51 datasets.

Table 1  Performance comparison of different approaches employing temporal features exclusively, on each of the four datasets

| Dataset | Method | Description | Acc (%) |
| --- | --- | --- | --- |
| CCV (Jiang et al. 2011) | Wu et al. (2015) | Temporal LSTM | 54.70 |
|  | Wu et al. (2015) | Temporal ConvNet | 59.10 |
|  | Li et al. (2020) | Temporal ResNet + hierarchical attentions | 63.62 |
|  | Zhang and Xiang (2020) | Temporal LSTM to *fc8* of VGG-19 | 63.80 |
|  | **Ours** (Jana et al. 2019) | **Temporal action localization, CNN + LSTM** | **79.13** |
| KCV (Loui et al. 2007) | Duan et al. (2012) | SIFT features | 35.46 |
|  | Chen et al. (2013) | Space-time features + MDA-HS | 49.61 |
|  | Luo et al. (2018) | Semi-supervised feature analysis | 47.70 |
|  | **Ours** (Jana et al. 2019) | **Temporal action localization, CNN+LSTM** | **52.41** |
| UCF-101 (Soomro et al. 2012) | Wu et al. (2015) | Temporal LSTM | 76.60 |
|  | Wu et al. (2015) | Temporal ConvNet | 78.30 |
|  | Wang et al. (2016) | ConvNets (Optical-flow + warped flow) | 87.80 |
|  | Mazari and Sahbi (2019) | Temporal pyramid + multiple representation | 68.58 |
|  | Peng et al. (2017) | Temporal stream of 2-network VGG-19 | 78.22 |
|  | Zang et al. (2018) | Temporal-weighted CNN + attention | 88.30 |
|  | Zhang and Xiang (2020) | Temporal GRU to *fc8* of VGG-19 | 64.50 |
|  | **Ours** (Jana et al. 2019) | **Key-frame, temporal CNN + LSTM** | **66.57** |
| HMDB-51 (Kuehne et al. 2011) | Girdhar et al. (2017) | ActionVLAD flow-stream | 59.10 |
|  | Cherian and Gould (2019) | ResNet-152 (Frame-level SMAID + Opt-Flow) | 59.50 |
|  | Zhang and Xiang (2020) | Temporal LSTM to *fc8* of VGG-19 | 33.30 |
|  | Li et al. (2020) | Temporal ResNet + hierarchical attentions | 37.29 |
|  | **Ours** (Jana et al. 2019) | **Temporal action localization, CNN + LSTM** | **55.67** |

Evaluation is made based on average accuracy (Acc) over all classes, expressed in percentage

The second stage evaluation is based upon improving this deficiency, with an efficient multimodality fusion strategy. We made this experiment with only two multimodalities viz., spatial and temporal. But this can be anytime extended to include other modalities too, since our fusion strategy is independent of the input modality. To be specific, we employ a late decision-fusion strategy. This goes on aggregating frame-level and video-level predictions obtained from CNN and LSTM respectively of different modalities, until there remains a single prediction vector. This final probability vector acts as the prediction made by the CNN-LSTM architecture. To evaluate our fusion strategy, in Table 2, we have tabulated some of the recent techniques that employ multimodality fusion. While it is seen that most researchers (Wu et al., 2015; Li et al., 2020; Wang et al., 2016; Cherian & Gould, 2019, etc.) prefer to use the average fusion, few (Zhang & Xiang, 2020; Mazari & Sahbi, 2019) prefer other variants. It is certainly debatable and reliant upon a test of time, that which is the most effective and robust fusion strategy. But we have compared the efficacy of the various fusion methods, by how much it could increment the accuracy of individual streams. It is visible from Table 2 that, our fusion strategy could increment the highest individual modality score by 15.31%, 8.53%, 7.01% and 6.69% respectively on CCV, KCV, UCF-101 and HMDB-51 datasets. This %age improvement in accuracy surpasses all the tabulated SoTA methods in the first three datasets, but assumes a second position for HMDB-51 dataset.

Overall, we can conclude that an effective event and activity recognition method that drives a video surveillance system should possess an efficient temporal feature exploitation and an unparalleled multimodality fusion scheme. Both of these are necessary for a Cyber-Physical System (CPS) dedicated towards video surveillance.

## 5  Epilogue and Way Forward

A detailed insight into the problem of classifying activities and events for Cyber-Physical Systems (CPS) is provided in this work. From inception days to present-day advances, we talk at length about each the approaches that have grown over time, contemporarily forming the base of the solution framework on various grounds, their subsequent drawbacks and how the successive frameworks have made efforts to eradicate them as much as possible. We exhibit the prowess of solely using temporal motion features to characterize event classes, keeping aside the prejudiced notion of CNNs gracing the spatial information better. To effectively demonstrate the same, we resort to a hybrid model of a ResNet50-LSTM fusion. This considerably surpasses the state-of-the art methods exploit-

Table 2 Performance comparison of different approaches employing multimodality fusion, on each of the four datasets

| Dataset | Method | Description | Fusion | Acc (%) |
|---|---|---|---|---|
| CCV (Jiang et al. 2011) | Wu et al. (2015) | ConvNet (spatial + temporal) | Average | 75.80 (S + 00.80) |
| | Wu et al. (2015) | ConvNet + LSTM (spatial + temporal) | Average | 81.70 (S + 03.80) |
| | Wu et al. (2015) | ConvNet + LSTM (spatial + temporal + acoustic) | Average | 82.40 (S + 04.50) |
| | Li et al. (2020) | Spatio-temporal ResNet + hierarchical Attn | Average | 74.21 (S + 08.14) |
| | Zhang and Xiang (2020) | GRU on transferred CNN | End-to-end network | 75.10 (N/A) |
| | **Ours (Jana et al. 2019)** | **Key-Frame, Spatio-Temporal CNN+LSTM** | **Multi-tier conflation** | **81.89 (S + 15.31)** |
| KCV (Loui et al. 2007) | Wang et al. (2016) | HDCC + SIFT | Group-weighing | 34.69 (N/A) |
| | Feng et al. (2014) | Multi-group adaptation n/w | – | 44.70 (N/A) |
| | **Ours (Jana et al. 2019)** | **Key-frame, Spatio-Temporal CNN + LSTM** | **Multi-tier conflation** | **57.52 (T + 08.53)** |
| UCF-101 (Soomro et al. 2012) | Wu et al. (2015) | ConvNet (spatial + temporal) | Average | 86.20 (S + 03.60) |
| | Wu et al. (2015) | ConvNet+LSTM (spatial + temporal) | Average | 90.10 (S + 06.10) |
| | Wu et al. (2015) | ConvNet+LSTM (spatial + temporal + acoustic) | Average | 90.30 (S + 06.30) |
| | Wang et al. (2016) | Spatio-Temporal ConvNets (temporal segment n/w) | Average | 93.50 (T + 05.60) |
| | Mazari and Sahbi (2019) | 3-D 2-stream (combined) + Temp. pyramid | Hierarchical aggregation | 97.94 (T + 01.53) |
| | Peng et al. (2017) | Spatial ResNet152 + temporal ResNet50 | Average | 87.80 (S + 06.70) |
| | Zang et al. (2018) | Spatio-temporal ConvNet | Attention model | 94.60 (T + 06.30) |
| | Zhang and Xiang (2020) | GRU on transferred CNN | End-to-end network | 88.00 (N/A) |
| | **Ours (Jana et al. 2019)** | **Key-frame, Spatio-temporal CNN + LSTM** | **Multi-tier conflation** | **89.03 (S + 07.01)** |
| HMDB-51 (Kuehne et al. 2011) | Girdhar et al. (2017) | ActionVLAD (RGB + Flow) Streams | ActionVLAD Late-Fuse | 66.90 (T + 07.80) |
| | Cherian and Gould (2019) | ResNet-152 (RGB + Opt-Flow + SMAID) | Average | 63.50 (T + 04.00) |
| | Zhang and Xiang (2020) | GRU on transferred CNN | End-to-end network | 59.10 (N/A) |
| | Li et al. (2020) | Spatio-temporal ResNet + Hierarchical Attn | Average | 70.69 (S + 02.31) |
| | **Ours (Jana et al. 2019)** | **Key-frame, spatio-temporal CNN + LSTM** | **Multi-tier conflation** | **61.91 (S + 06.69)** |

Evaluation is made based on average accuracy (Acc) over all classes, expressed in percentage. In the rightmost column, value in bracket indicates increment (+)/decrement (−) from the highest individual performance of a modality stream. S = spatial, T = temporal, A = acoustic ing only action features, and is found to be comparable to some which make use of the visual cues offered by the frame images. Our multi-level fusion strategy uses the concept of biased conflation and takes into account verdicts from both the spatial and temporal wings of the network, as well as on the grounds of frame and video level predictions. It can be said that the robustness of this novel statistical method lies in the fact that it has the potential to effectively "fuse" two apparently wrong or disagreeing predictions obtained from the respective spatial and temporal wings to an eventual correct classification. This is observed as it undergoes various formative stages in the method of fusion. It may greatly help in practical scenarios of doubt in surveillance for a CPS. For example hypothetically, while the spatial content for a mali-cious intruder in an ATM booth would not vary that much as compared to a general customer, suspicious activities other than what is usually expected can be recognized by the temporal wing. The confidence of verdicts of both of these can then undergo the fusion process for the decision to be made by the model (so as to fire an alarm on detected intrusion or not). Apart from these merits, the model could be improved in dealing with classes having a high degree of semantic correlation (for example, Boxing and Punching, Soccer Juggling and Kicking). Modifications in the network model to account for such scenarios could be made. Also, we are yet to exploit auditory information in classifying these events. Finally, guiding the CNN or RNN model through an effective 'attention' scheme could add clarity to the way each activity class is characterized. This could be explored, along with constructing a suitable and generic interpretation for what could be exploited in this 'fovea' or 'attention' stream.